\documentclass[conference]{IEEEtran}
\IEEEoverridecommandlockouts
\usepackage{cite}
\usepackage{amsmath,amssymb,amsfonts}
\usepackage{algorithmic}
\usepackage{graphicx}
\usepackage{textcomp}
\usepackage{xcolor}
\def\BibTeX{{\rm B\kern-.05em{\sc i\kern-.025em b}\kern-.08em
    T\kern-.1667em\lower.7ex\hbox{E}\kern-.125emX}}
\begin{document}

\title{Algorithms for Object Detection in Substations}

\author{\IEEEauthorblockN{Bingying Jin}
\IEEEauthorblockA{\textit{Academy of Electrical Engineering} \\
\textit{Shanghai Jiaotong University}\\
Shanghai, China}
\and
\IEEEauthorblockN{Yadong Liu}
\IEEEauthorblockA{\textit{Academy of Electrical Engineering} \\
\textit{Shanghai Jiaotong University}\\
Shanghai, China}
\and
\IEEEauthorblockN{Qinlin Qian}
\IEEEauthorblockA{\textit{Academy of Electrical Engineering} \\
\textit{Shanghai Jiaotong University}\\
Shanghai, China}
}

\maketitle

\begin{abstract}
Inspection of high-voltage power equipment is an effective way to ensure power supply reliability. Object recognition, one of the key technologies in automatic power equipment inspection, attracts attention of many researchers and engineers. Although quite a few existing models have some their own advantages, object relationship between equipment which is very important in this task is scarcely considered. This paper combining object relationship modeling and Transformer Model proposes a Relation Transformer Model. It has four parts--- backbone, encoder, decoder and prediction heads. With this structure, the proposed method shows in experiments a much better performance than other three commonly used models in object recognition in substation, largely promoting the development of automatic power equipment inspection. 
\end{abstract}


\section{Introduction}
High-voltage equipment inspection is an effective way to ensure the power supply reliability. With the implementation of fixed-position policy in state-owned enterprises and the development of the power grid, the problems of equipment inspection work have become increasingly prominent, mainly in:

1) With fewer people and more tasks, the inspection task is difficult to perform, and the inspection quality is low. At present, the main operation and inspection method of converter stations/substations of power grid companies with voltage levels above 500kV is manual inspection. The inspection workload accounts for more than 50\% of the operating personnel. The manual inspection method has single functions and unstable inspection quality, resulting in an overall low inspection quality.

2) The comprehensiveness and intelligence of equipment inspection need to be improved urgently. Due to the limitation of the inspection route, there are a large number of blind spots in the inspection range. At the same time, only infrared and visible light functions are integrated. The level of intelligent analysis is not good. Basically, manual reading is the main method for image analysis. There is no inspection of the insulation state, which leads to the problem of low comprehensiveness.

If advanced detection and diagnosis technology can be used to realize unmanned inspection of key equipment in substations and automatic judgment of hidden dangers and abnormal conditions, it can greatly liberate the inspection burden of front-line crews and help the construction of front-line teams change into management work\cite{jenssen2019intelligent,wang2019research,guo2010mobile}. The management transformation has greatly improved the current substation operation and maintenance level.

The research and application of image processing in the state detection of substation equipment are mainly concentrated in the aspects of transmission line aircraft inspection, sag measurement and infrared and ultraviolet image analysis\cite{velasquez2016robot, shengfang2006research}. The main work of image detection and analysis of substation equipment is to improve the efficiency and accuracy of the flight inspection and image monitoring of transmission lines. The main research points are concentrated on the extraction and identification of transmission lines, towers and insulators, as well as the identification of specific defects or states such as icing, broken strands, and sagging. There are merely preliminary research results. A small amount of image monitoring equipment such as sag image measurement has formed application products. Judging from the number of documents reviewed, there is not much research work on image processing in the state detection of substation equipment, and the image processing technology used is not cutting-edge, and the image characteristics of substation equipment are less involved. No image analysis system software product for fault diagnosis of various substation equipment has been developed.

This paper is to solve the key problem in high-voltage equipment inspection---object recognition in substation. A novel model—Relation Transformers, modified from the model of Transformers, is proposed to largely improve the accuracy of this task with the help of relationship between power equipment. The network structure contains four parts: backbone, encoder, decoder and prediction heads, and object relationship modeling is added in the decoder module.

\section{NETWORK STRUCTURE}

\subsection{Network Architecture}

In Fig. 1, network architecture of Relation Transformers consisting of four parts - backbone, encoder, decoder, and prediction heads \cite{carion2020end}, is shown. Convolutional Neural Network(ResNet50\cite{he2016deep}) is adopted as the backbone network which extracts feature maps $f \in R^{C \times H \times W}$ from original pictures. Then the channel dimension of $f$ is reduced by a $1 \times 1$ convolution layer from $C$ to $d$. After that, a new feature map $z_0 \in R^{d \times H \times W}$ is created, the spatial dimensions of which are then squeezed into one dimension, resulting in the input of encoder of transformer $z_o^{\prime} \in R^{d \times H W}$. There are two parts in a standard encoder structure: a multi-head selfattention \cite{vaswani2017attention} module and a feed forward network. Here fixed positional encodings \cite{bello2019attention, parmar2018image} are added to the input of attention layer to make sure that the architecture is invariant of permutation.

To consider the relationship between objects, a graph is constructed to find other related objects of a certain object \cite{xiong2021object, siheng2020power}. In this graph, each object is a node connected with its neighbors. When there is an edge between two objects, they are neighbors and have a strong relationship. Here some explanations about object relationship are given to make it easier to understand. Not all the relationship between every two objects is considered since: 1. The complexity of the network should be controlled to make the process of learning and prediction efficient; 2. Not all the relationships are equally important. The importance of relationship is determined by the distance between these two objects. Focusing on stronger relationship is more helpful when learning with few labeled samples.

The features of an object are from itself and its neighbors, which is implemented with a neural network layer. Then the fixed features are used as input of the standard decoder \cite{dosovitskiy2020image, zhu2020deformable}, which using multi-head self-attention modules and encoderdecoder attention modules transforms $N$ embeddings of size $d$. The $N$ different input embeddings are used to produce results, ensuring a permutation-invariant decoder. Object queries, namely learnt positional encodings, are added to the input ' of attention' layer, ' which are processed by the decoder' into output embeddings. The $N$ final predictions composed of box coordinates and class labels are independently generated by a feed forward network. In these embeddings, 'self-attention and encoder-decoder attention modules are adopted, providing the model pair-wise relationships and the whole context of the image, which largely improves the accuracy of detection.

The feed-forward network(FFN), which is used for final prediction, is a multi-layer perceptron followed by a linear layer. The multi-layer perceptron with a hidden dimension of $d$ uses ReLU activation. It predicts the position, height, width of the bounding box and the linear layer is a softmax function giving the class label. Here the background class is used to represent low-quality bounding boxes.

\section{Loss Function Design}

The central idea is to match the prediction and ground truth in a bipartite way, and then to calculate the specific class and bounding box losses. The set of ground truth is denoted by $y$, and the set of predictions is $\hat{y}=\left\{\hat{y}_i\right\}_{i=1}^N$. A bipartite matching between them can be represented as an optimal permutation $\sigma$. with the lowest cost:
$$
\hat{\sigma}=\underset{\sigma}{\arg \min } \sum_i^N L_{\operatorname{math}}\left(y_i, \hat{y}_{\sigma(i)}\right)
$$
Where $L_{\text {match }}\left(y_i, \hat{y}_{\sigma(i)}\right)$ is the cost of a matching between  $y$. and $\hat{y}$ with index $\sigma(i)$. Many algorithms such as Hungarian. algorithm \cite{kuhn1955hungarian} can be used to obtain the optimal assignment.
Class and bounding box of a predicted object are both considered in the pair-wise  matching  cost. Here ground truth $y_i=\left(c_i, b_i\right)$ where $c_i$ is the target of class and $b_i$ is a position vector  consisting bounding boxs  coordinates  and sizes. With these definition, it can be written as:
$$
L_{\text {match }}\left(y_i, \hat{y}_{\sigma(i)}\right)=-1_{\left\{c_i \neq \varnothing\right\}} \hat{p}_{\sigma(i)}\left(c_i\right)+1_{\left\{c_i \neq \varnothing\right\}} L_{\text {box }}\left(b_i, \hat{b}_{\sigma(i)}\right)
$$
Where $\hat{p}_{\sigma(i)}\left(c_i\right)$ is the probability of $c_i, \hat{b}_{\sigma(i)}$ is the prediction box with index $\sigma(i), \varnothing$ is the class of background.

With the Hungarian algorithm, Eq.(2) can be rewritten as:
$$
L_{H-match}(y, \hat{y})=\sum_{i=1}^N\left[-\log \hat{p}_{\tilde{\sigma}(i)}\left(c_i\right)+1_{\left\{c_i \neq \varnothing\right\}} L_{\text {box }}\left(b_i, \hat{b}_{\hat{\sigma}(i)}\right)\right]
$$
Where $\hat{\sigma}$ is the assignment after optimization.
Finally, it is discussed below how to calculate the bounding box loss. To eliminate the impact of scale, $L_1$ loss and thegeneralized- IoU- loss  $L_{\text {iou }}$ \cite{rezatofighi2019generalized} - are linearly combined as the bounding box loss, that is:
$$
L_{\text {box }}\left(b_i, \hat{b}_{\sigma(i)}\right)=\lambda_{\text {iou }} L_{\text {iou }}\left(b_i, \hat{b}_{\sigma(i)}\right)+\lambda_1\left\|b_i-\hat{b}_{\sigma(i)}\right\|_1
$$
Where $\lambda_1$ and $\lambda_{\text {Iov }}$ are corresponding weights.

\section{EXPERIMENTS}

In this section, the proposed method, Faster R-CNN[17], YOLO[18] and SSD[19] are compared with the images of power equipment in substation.

\subsection{Dataset}
The dataset contains 1000 pictures of power equipment in substation, and 800 of them are taken by inspection robots and 200 of them are from the Internet. All the pictures are checked and annotated by human. There are two types of objects: power equipment and foreign objects. For power equipment, there are transformer, GIS, insulator, arrester, PT, CT, switch, connecting head, disconnector, bus, bushing. For foreign objects, there are inspection robots and unmanned aerial vehicle(UAV). An illustration of typical scenes in the dataset is shown in Fig. 3. 

\subsection{Results of proposed method}
The results of proposed method are given in Fig.4. In Fig.4, all the detected objects are with a bounding box representing its position and size, the color of which denotes different classes. For example, the red color bounding box represents a transformer, the green color shows the UAV, the blue color indicates inspection robots, the yellow color denotes bushing, and the purple color means insulator. It can be seen that the proposed method has a very high accuracy over all kinds of objects including both power equipment and foreign objects.

\subsection{Comparison with other models}

The mainstream algorithms of target detection are mainly divided into two types: (1) Two-stage methods, such as R-CNN algorithms[20], whose main idea is to use heuristic methods (Selective search[21]) or CNN network (RPN) first, generates a series of sparse candidate boxes, and then classifies and regresses these candidate boxes. The advantage of the two-stage method is high accuracy; (2) One-stage methods, such as Yolo and SSD. The main idea of them is to evenly conduct dense sampling at different positions of the picture. When sampling, different scales and aspect ratios can be used. Then classification and regression are directly performed based on the extracted features by CNN. It is faster than two-stage methods at the cost of more difficult training process due to dense sampling. This is mainly because the positive sample and the negative sample (background) are extremely imbalanced, resulting in slightly lower model accuracy.

Faster R-CNN, modified from Fast R-CNN[22], integrates feature extraction, proposal extraction, bounding box regression, and classification into one network, which greatly improves the overall performance, especially in terms of detection speed. It has four parts: 1. Convolutional layers are used to extract the features from original pictures. The input is the entire picture, and the output is the extracted features. The feature maps are shared for use in subsequent RPN networks and fully connected layers. 2. The RPN network is used to generate region proposals. 3. ROI Pooling. After integrating the input feature maps and proposals, this layer extracts the proposal feature maps to obtain a fixed-size feature map, and then sends it to the subsequent fully connected layer to determine the target category. 4. Classification and regression, the output of this layer is the final goal: the class of the candidate area and its precise location.

YOLO(You Only Look Once) is an object recognition and localization algorithm based on deep neural network. Its biggest four features are: 1. One-stage. 2. Predicting multiple bounding boxes and categories at the same time. 3. End-to-end target detection and recognition. 4. Faster: The CNN that realizes the regression function does not need complicated design. It directly selects the whole image to train the model in order to better distinguish the target and background area.

SSD(Single Shot Multi-Box Detector) is one of the main detection frameworks at present. Based on the idea of converting detection into regression, it can complete target positioning and classification at the same time. The algorithm is based on the Anchor technology in Faster R-CNN, proposing a similar Prior box. The algorithm modifies the traditional VGG16 network[23]: converts the FC6 and FC7 layers of VGG16 into convolutional layers Conv6 and Conv7; removes all Dropout layer and FC8 layer; adds the Atrous algorithm to transform Pool5 from 2x2-S2 to 3x3-S1, and at the same time adds the detection layer based on the feature pyramids, that is, the feature maps of different receptive fields.
Although these three popular object detection algorithms are advanced and have their own advantages, they do not consider object relationship, a kind of very important information in the scene of substation, which impairs their ability to learn with few examples and complicated situations. Therefore the proposed model which make full use of object relation performs much better than them as shown in Table I.

The metrics used here is Average Precision [24]. In pattern recognition, information retrieval and classification, precision is the fraction of relevant instances among the retrieved instances, while recall is the fraction of relevant instances that were retrieved. Obviously, the higher precision, the lower recall and vice versa. And Average Precision is the average of all precision value under each recall value. It can be concluded from Table I that the proposed model has a much better performance than other three models, the reason of which is that the proposed model considers object relationship which helps it learn and understand with few samples and complicated situations. Besides, it has a high accuracy over all classes, meaning that this method applies to different equipment and therefore has a good generalization quality. 

\section{CONCLUSION}

This paper combining object relationship modeling and Transformer Model proposes a Relation Transformer Model which has four parts---backbone, encoder, decoder and prediction heads. The backbone network which generates feature map from original images adopts Convolutional Neural Network. Then features are sent into encoder and decoder for self-attention-mechanism-based adjustment and fusion. Here object relation modeling is considered in decoder. Finally, feed-forward network is used to predict class label and the position, height, width of the bounding box.

With this structure, the proposed method shows in experiments a much better performance than other three commonly used models in object recognition in substation. In addition to a higher accuracy, the proposed model has a higher calculation efficiency, extensibility, flexibility and feature extraction ability. Therefore, this model largely promotes the development of automatic power equipment inspection.

\bibliographystyle{IEEEtran}
\bibliography{ieee_conference}

\end{document}